\documentclass{sig-alternate-05-2015}

\usepackage{amsmath}
\usepackage{multirow}
\usepackage{url}
\usepackage{graphicx}
\usepackage{hyperref}

\begin{document}

%\setcopyright{acmcopyright}

\title{Contextual LSTM (CLSTM) models for \\ Large scale NLP tasks}
%\title{Contextual LSTM models}
\numberofauthors{6}

\author{
  \alignauthor Shalini Ghosh\titlenote{Work was done while visiting Google Research.} \\ \email{shalini@csl.sri.com}
  \alignauthor Oriol Vinyals \\ \email{vinyals@google.com}
  \alignauthor Brian Strope \\ \email{bps@google.com} 
\and
  \alignauthor Scott Roy \\ \email{hsr@google.com} 
  \alignauthor Tom Dean \\ \email{tld@google.com} 
  \alignauthor Larry Heck \\ \email{larryheck@google.com}
}
\date{}

\maketitle

\begin{abstract}
Documents exhibit sequential structure at multiple levels of
abstraction (e.g., sentences, paragraphs, sections). These
abstractions constitute a natural hierarchy for representing the
context in which to infer the meaning of words and larger fragments of
text. In this paper, we present CLSTM (Contextual LSTM), an extension
of the recurrent neural network LSTM (Long-Short Term Memory) model,
where we incorporate contextual features (e.g., topics) into the
model. We evaluate CLSTM on three specific NLP tasks: word prediction,
next sentence selection, and sentence topic prediction. Results from
experiments run on two corpora, English documents in Wikipedia and a
subset of articles from a recent snapshot of English Google News,
indicate that using both words and topics as features improves
performance of the CLSTM models over baseline LSTM models for these
tasks. For example on the next sentence selection task, we get
relative accuracy improvements of 21\% for the Wikipedia dataset and
18\% for the Google News dataset. This clearly demonstrates the
significant benefit of using context appropriately in natural language
(NL) tasks. This has implications for a wide variety of NL
applications like question answering, sentence completion, paraphrase
generation, and next utterance prediction in dialog systems.
\end{abstract}

\section{Introduction}
\label{sec:intro}

Documents have sequential structure at different hierarchical levels
of abstraction: a document is typically composed of a sequence of
sections that have a sequence of paragraphs, a paragraph is
essentially a sequence of sentences, each sentence has sequences of
phrases that are comprised of a sequence of words, etc. Capturing this
hierarchical sequential structure in a language model
(LM)~\cite{ManningandSchutze99} can potentially give the model more
predictive accuracy, as we have seen in previous
work~\cite{HihiandBengioNIPS-96,FernandezetalIJCAI-07,KurzweilBook,MnihandHintonNIPS-08,ZhuetalCoRR-15}.

A useful aspect of text that can be utilized to improve the
performance of LMs is long-range context. For example, let us 
consider the following three text segments:

1) Sir Ahmed Salman Rushdie is a British Indian novelist and
  essayist. He is said to combine {\em magical} realism with
  historical fiction.

2) Calvin Harris \& HAIM combine their powers for a {\em magical}
  music video.

3) Herbs have enormous {\em magical} power, as they hold the earth's
  energy within them.

Consider an LM that is trained on a dataset having the example
sentences given above --- given the word ``magical'', what should be
the most likely next word according to the LM: realism, music, or
power? In this example, that would depend on the longer-range context
of the segment in which the word ``magical'' occurs. One way in which
the context can be captured succinctly is by using the topic of the
text segment (e.g., topic of the sentence, paragraph). If the context
has the topic ``literature'', the most likely next word should be
``realism''.  This observation motivated us to explore the use of
topics of text segments to capture hierarchical and long-range context
of text in LMs.

In this paper, we consider Long-Short Term Memory (LSTM)
models~\cite{hochreiter}, a specific kind of Recurrent Neural Networks
(RNNs). The LSTM model and its different variants have achieved
impressive performance in different sequence learning problems in
speech, image, music and text
analysis~\cite{GersetalJMLR-02,GravesetalASRU-13,GravesandSchmidhuberIJCNN-05,SaketalINTERSPEECH-14a,SutskeverPhD-13,SutskeveretalCoRR-14,vinyalsetal,vinyalsetalcvpr,xu:caption2015},
where it is useful in capturing long-range dependencies in sequences.
LSTMs substantially improve our ability to handle long-range
dependencies, though they still have some limitations in this
regard~\cite{ChoetalCoRR-14,HihiandBengioNIPS-96}.
% The power of LSTMs lies in being able to remember states for a long time.  

RNN-based language models (RNN-LMs) were proposed by Mikolov et
al.~\cite{MikolovKBCK10}, and in particular the variant using LSTMs
was introduced by Sundermeyer et al.~\cite{SundermeyerSN12}. In this
paper, we work with LSTM-based LMs. Typically LSTMs used for language
modeling consider only words as features. Mikolov et
al.~\cite{MikolovandZweigSLT-12} proposed a conditional RNN-LM for
adding context --- we extend this approach of using context in RNN-LMs
to LSTMs, train the LSTM models on large-scale data, and propose new
tasks beyond next work prediction.

We incorporate contextual features (namely, topics based on different
segments of text) into the LSTM model, and call the resulting model
Contextual LSTM (CLSTM). In this work we evaluate how adding
contextual features in the CLSTM improves the following tasks:

1) {\em Word prediction}: Given the words and topic seen so far in
the current sentence, predict the most likely next word. This task is
important for sentence completion in applications like {\em predictive
  keyboard}, where long-range context can improve word/phrase
prediction during text entry on a mobile phone.

2) {\em Next sentence selection}: Given a sequence of sentences,
find the most likely next sentence from a set of candidates. This is
an important task in {\em question/answering}, where topic can be
useful in selecting the best answer from a set of template
answers. This task is also relevant in other applications like Smart
Reply~\cite{smart_reply}, for predicting the best response to an email
from a set of candidate responses.

3) {\em Sentence topic prediction}: Given the words and topic of
the current sentence, predict the topic of the next sentence. We
consider two scenarios: (a) where we don't know the words of the next
sentence, (b) where we know the words of the next sentence. Scenario
(a) is relevant for applications where we don't know the words of a
user's next utterance, e.g., while predicting the {\em topic of
  response of the user of a dialog system}, which is useful in knowing
the intent of the user; in scenario (b) we try to predict the
topic/intent of an utterance, which is common in a {\em topic
  modeling} task.\\

\noindent The main contributions of this paper are as follows:

1) We propose a new Contextual LSTM (CLSTM) model, and
demonstrate how it can be useful in tasks like word prediction, next
sentence scoring and sentence topic prediction -- our experiments show
that incorporating context into an LSTM model (via the CLSTM) gives
improvements compared to a baseline LSTM model. This can have
potential impact for a wide variety of NLP applications where these
tasks are relevant, e.g. sentence completion, question/answering,
paraphrase generation, dialog systems.

2) We trained the CLSTM (and the corresponding baseline
LSTM) models on two large-scale document corpora: English documents
in Wikipedia, and a recent snapshot of English Google News documents.
The vocabulary we handled in the modeling here was also large: 130K
words for Wikipedia, 100K for Google news. Our experiments and
analysis demonstrate that the CLSTM model that combines the power of
topics with word-level features yields significant performance gains
over a strong baseline LSTM model that uses only word-level
features. For example, in the next sentence selection task, CLSTM gets
a performance improvement of 21\% and 18\% respectively over the LSTM
model on the English Wikipedia and Google News datasets.

3) We show initial promising results with a model where we
learn the thought embedding in an unsupervised manner through the
model structure, instead of using supervised extraneous topic as side
information (details in Section~\ref{sec:unsup}).

%For instance, on the task of finding the best-scoring next sentence
%from a set of candidate sentences, CLSTM gets relative accuracy
%improvements of 20\% on the Wikipedia dataset and 18\% on the Google
%News subset, over the corresponding baseline LSTM model.

\section{Related Work}
\label{sec:related}

There are various approaches that try to fit a generative model for
full documents. These include models that capture the content
structure using Hidden Markov Models (HMMs)~\cite{BarzilayHLT04}, or
semantic parsing techniques to identify the underlying meanings in
text segments~\cite{LuEMNLP08}.  Hierarchical models have been
used successfully in many applications, including hierarchical
Bayesian models~\cite{DeanAMAI-06,LeeandMumfordJOSA-03}, hierarchical
probabilistic models~\cite{SocheretalISBI-08}, hierarchical
HMMs~\cite{FineetalML-98} and hierarchical
CRFs~\cite{ReynoldsandMurphyCRV-07}.

As mentioned in Section~\ref{sec:intro}, RNN-based language models
(RNN-LMs) were proposed by Mikolov et al.~\cite{MikolovKBCK10}, and
the variant using LSTMs was introduced by Sundermeyer et
al.~\cite{SundermeyerSN12} -- in this paper, we work with LSTM-based
LMs. Mikolov et al.~\cite{MikolovandZweigSLT-12} proposed a
conditional RNN-LM for adding context --- we extend this approach of
using context in RNN-LMs to LSTMs.

Recent advances in deep learning can model hierarchical structure
using deep belief
networks~\cite{Huang:2013,ZhuetalCoRR-15,ZorziPsy15,baidu:2016},
especially using a hierarchical recurrent neural network (RNN)
framework.  In Clockwork RNNs~\cite{KoutniketalICML-14} the hidden
layer is partitioned into separate modules, each processing inputs at
its own individual temporal granularity. Connectionist Temporal
Classification or CTC~\cite{GravesetalCoRR-13} does not explicitly
segment the input in the hidden layer -- it instead uses a
forward-backward algorithm to sum over all possible segments, and
determines the normalized probability of the target sequence given the
input sequence. Other approaches include a hybrid NN-HMM
model~\cite{AbdelHamidMJP12}, where the temporal dependency is handled
by an HMM and the dependency between adjacent frames is handled by a
neural net (NN).  In this model, each node of the convolutional hidden
layer corresponds to a higher-level feature.

Some NN models have also used context for modeling text. Paragraph
vectors~\cite{dai2014document,LeandMikolovCoRR-14} propose an
unsupervised algorithm that learns a latent variable from a sample of
words from the context of a word, and uses the learned latent context
representation as an auxiliary input to an underlying skip-gram or
Continuous Bag-of-words (CBOW) model. Another model that uses the
context of a word infers the Latent Dirichlet Allocation (LDA) topics
of the context before a word and uses those to modify a RNN model
predicting the word~\cite{MikolovandZweigSLT-12}.

Tree-structured LSTMs~\cite{TaietalCoRR-15,ZhuetalCoRR-15} extend
chain-structured LSTMs to the tree structure and propose a principled
approach of considering long-distance interaction over hierarchies,
e.g., language or image parse structures. Convolution networks have
been used for multi-level text understanding, starting from
character-level inputs all the way to abstract text
concepts~\cite{ZhangL15}.  Skip thought vectors have also been used to
train an encoder-decoder model that tries to reconstruct the
surrounding sentences of an encoded passage~\cite{Kiros15}.

Other related work include Document Context Language
models~\cite{ji_document_context}, where the authors have multi-level
recurrent neural network language models that incorporate context from
within a sentence and from previous sentences. Lin et
al.~\cite{lin_emnlp_2015} use a hierarchical RNN structure for
document-level as well as sentence-level modeling -- they evaluate
their models using word prediction perplexity, as well as an
approach of coherence evaluation by trying to predict 
sentence-level ordering in a document. 

In this work, we explore the use of long-range hierarchical signals
(e.g., sentence level or paragraph level topic) for text analysis
using a LSTM-based sequence model, on large-scale data --- to the best
of our knowledge this kind of contextual LSTM models, which model the
context using a 2-level LSTM architecture, have not been trained
before at scale on text data for the NLP tasks mentioned in
Section~\ref{sec:intro}.

%% \section{Background}
%% \label{sec:back}

%% \begin{figure}[hbtp]
%%   \includegraphics[width=\columnwidth]{figs/fig-1.pdf}
%%   \caption{Unrolled view of RNN models, e.g., LSTM}
%%   \label{fig:lstm-model}
%% \end{figure}

%% Figure~\ref{fig:lstm-model} is a
%% schematic diagram showing how the LSTM model generates a sequence of
%% outputs given a sequence of inputs, where the recurrent structure of
%% the LSTM is unrolled in time to show how a sequence is input and
%% output from the model. 
%(Figure~\ref{fig:lstm-cell} shows a basic LSTM cell). The 

%% The following equations represent the operations of the different
%% components of the LSTM cell~\cite{GravesPhD-09}\footnote{We present
%%   the LSTMs equations over here, since we will show subsequently how
%%   we have modified these equations to incorporate topics into the LSTM
%%   cells.}:

%% \begin{eqnarray}
%% i_t &=&  \sigma(W_{xi} x_t + W_{hi} h_{t-1} + W_{ci} c_{t-1} + b_i)  \nonumber \\
%% f_t &=&  \sigma(W_{xf} x_t + W_{hf} h_{t-1} + W_{cf} c_{t-1} + b_f) \nonumber \\
%% c_t &=& f_t c_{t-1} + i_t \tanh (W_{xc} x_t + W_{hc} h_{t-1} + b_c) \nonumber \\
%% o_t &=&  \sigma(W_{xo} x_t + W_{ho} h_{t-1} + W_{co} c_{t} + b_o) \nonumber \\
%% h_t &=& o_t \tanh(c_t) \label{eq:lstm-h}
%% \end{eqnarray}

%\begin{figure}[hbtp]
%\centering
%  \includegraphics[width=1.2\columnwidth]{figs/fig-2.pdf}
%  \caption{LSTM cell (Figure adapted from: {\small \url{http://christianherta.de/lehre/dataScience/machineLearning/neuralNetworks/LSTM.php}})}
%  \label{fig:lstm-cell}
%\end{figure}

\section{Word Prediction}
\label{sec:wordpred}

Of the three different tasks outlined in Section~\ref{sec:intro},
we focus first on the word prediction task, where the goal is to
predict the next word in a sentence given the words and context
(captured via topic) seen previously.

Let $s_i$ be the $i^{th}$ sentence in a sequence of sentences,
$w_{i,j}$ be the $j^{th}$ word of sentence $s_i$, $n_i$ be the number
of words in $s_i$, and $w_{i,j} \ldots w_{i,k}$ indicate the sequence
of words from word $j$ to word $k$ in sentence $i$. Note that sentence
$s_i$ is equivalent to the sequence of words $w_{i,0} \ldots
w_{i,n_i-1}$. Let $T$ be the random variable denoting the topic -- it
is computed based on a particular subsequence of words seen from the
first word of the sequence ($w_{0,0}$) to the current word
($w_{i,j}$). This topic can be based on the current sentence segment
(i.e., $T = Topic(w_{i,0} \ldots w_{i,j-1})$), or the previous
sentence (i.e., $T = Topic(w_{i-1,0} \ldots w_{i-1,n_{i-1}})$),
etc. Details regarding the topic computation are outlined in
Section~\ref{sec:hitom}.

Using this notation, the word prediction task in our case can be
specified as follows: given a model with parameters $\Theta$, words
$w_{0,0} \ldots w_{i,j}$ and the topic $T$ computed from a
subsequence of the words from the beginning of the sequence, find the
next word $w_{i,j+1}$ that maximizes the probability:
$P(w_{i,j+1} | w_{0,0} \ldots w_{i,j}, T, \Theta)$.

\subsection{Model}
\label{sec:wordpred-model}

For our approach, as explained before, we introduce the power of
context into a standard LSTM model. LSTM is a recurrent neural network
that is useful for capturing long-range dependencies in sequences.
The LSTM model has multiple LSTM cells, where each LSTM cell models
the digital memory in a neural network.  It has gates that allow the
LSTM to store and access information over time. For example, the
input/output gates control cell input/output, while the forget gate
controls the state of the cell.

The word-prediction LSTM model was implemented in the large-scale
distributed Google Brain framework~\cite{DeanetalNIPS-12}. The model
takes words encoded in 1-hot encoding from the input, converts them to
an embedding vector, and consumes the word vectors one at a time. The
model is trained to predict the next word, given a sequence of words
already seen. The core algorithm used to train the LSTM parameters
is BPTT~\cite{werbos88}, using a softmax layer that uses the id of the
next word as the ground truth.

To adapt the LSTM cell that takes words to a CLSTM cell that takes as
input both words and topics, we modify the equations representing the
operations of the LSTM cell~\cite{GravesPhD-09} to add the topic
vector $T$ to the input gate, forget gate, cell and output gate ($T$
is the embedding of the discrete topic vector). In each of the
following equations, the term in bold is the modification made to the
original LSTM equation.
\begin{eqnarray}
i_t &=&  \sigma(W_{xi} x_t + W_{hi} h_{t-1} + W_{ci} c_{t-1} + b_i + \bf {W_{Ti} T)} \nonumber \\
f_t &=&  \sigma(W_{xf} x_t + W_{hf} h_{t-1} + W_{cf} c_{t-1} + b_f + \bf {W_{Ti} T)} \nonumber \\
c_t &=&  f_t c_{t-1} + i_t \tanh (W_{xc} x_t + W_{hc} h_{t-1} + b_c + \bf {W_{Ti} T)} \nonumber \\
o_t &=&  \sigma(W_{xo} x_t + W_{ho} h_{t-1} + W_{co} c_{t} + b_o + \bf {W_{Ti} T)} \nonumber \\
h_t &=&  o_t \tanh(c_t) \label{eq:clstm-o}
\end{eqnarray}

\begin{figure}[hbtp]
  \includegraphics[width=\columnwidth]{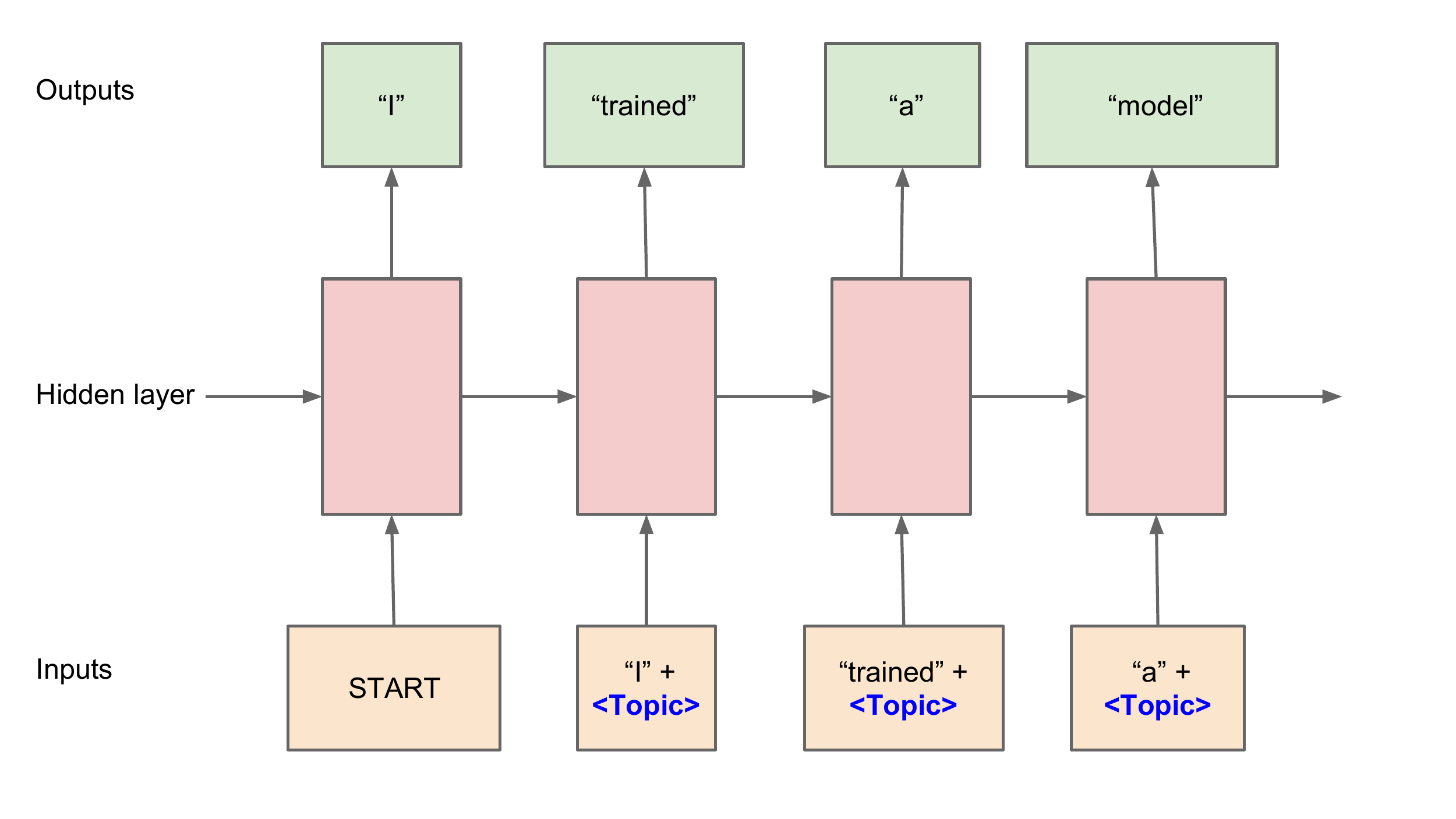}
  \caption{CLSTM model ($<$Topic$>$ = topic input)}
  \label{fig:hlstm-model}
\end{figure}

In these equations $i$, $f$ and $o$ are the input gate, forget gate
and output gate respectively, $x$ is the input, $b$ is the bias term,
$c$ is the cell memory, and $h$ is the output. As an example, consider
the input gate equation:
\begin{eqnarray}
i_t =&  \sigma(W_{xi} x_t + W_{hi} h_{t-1} + W_{ci} c_{t-1} + b_i) \nonumber \\
    =&  \sigma([W_{xi} \ W_{hi} \ W_{ci} \ 1] [x_t \ h_{t-1} \ c_{t-1} \ b_i]^T) \label{eq:lstm-a}
\end{eqnarray}
When we add the topic signal $T$ to the input gate, the equation is
modified to:
\begin{eqnarray}
i_t =&  \sigma(W_{xi} x_t + W_{hi} h_{t-1} + W_{ci} c_{t-1} + b_i + W_{Ti} T) \nonumber \\
    =&  \sigma([W_{xi} \ W_{Ti} \  W_{hi} \ W_{ci} \ 1] [x_t \ T \ h_{t-1} \ c_{t-1} \ b_i]^T) \label{eq:lstm-b}
\end{eqnarray}
Comparing the last two equations, Equations~\ref{eq:lstm-a}
and~\ref{eq:lstm-b}, we see that having a topic vector $T$ added into
the CLSTM cell is equivalent to considering a composite input $[x_i$
  \ $T]$ to the LSTM cell that concatenates the word embedding and
topic embedding vectors. This approach of concatenating topic and word
embeddings in the input worked better in practice than other
strategies for combining topics with words.
Figure~\ref{fig:hlstm-model} shows the schematic figure of a CLSTM
model that considers both word and topic input vectors.

Note that we add the topic input to each LSTM cell since each LSTM
cell can potentially have a different topic. For example, when the
topic is based on the sentence segment seen so far (see
Section~\ref{sec:features}), the topic is based on the current
sentence prefix --- so, each LSTM cell can potentially have a
different topic. Note that in some setups each LSTM cell in a layer
could have the same topic, e.g., when the topic is derived from
the words in the previous sentence.

%% Figure~\ref{fig:clstm-distbelief} shows the implementation of the
%% CLSTM model in the DistBelief framework, where the PSEmbedding Layer
%% maps the 1-hot encoded input to a dense encoding and is also learned
%% as part of the LSTM training~\cite{DeanetalNIPS-12}.

%% \begin{figure}[hbtp]
%%   \centering
%%   \includegraphics[width=1.5\columnwidth]{figs/LSTM_StateLayer_WordTopic.pdf}
%%   \caption{CLSTM implementation in DistBelief}
%%   \label{fig:clstm-distbelief}
%% \end{figure}

\subsection{HTM: Supervised Topic Labels}
\label{sec:hitom}

The topics of the text segments can be estimated using different
unsupervised methods (e.g., clustering) or supervised methods (e.g.,
hierarchical classification). For the word prediction task we use
HTM\footnote{Name of actual tool modified to HTM, abbreviation for
  \underline{Hi}erarchical \underline{To}pic \underline{M}odel, for
  confidentiality.}, a hierarchical topic model for supervised
classification of text into a hierarchy of topic categories, based on
the Google Rephil large-scale clustering tool~\cite{Murphy12}. There
are about 750 categories at the leaf level of the HTM topic
hierarchy. Given a segment of text, HTM gives a probability
distribution over the categories in the hierarchy, including both leaf
and intermediate categories. We currently choose highest probability
topic as the most-likely category of the text segment.

\subsection{Experiments}
\label{sec:wordpred-expt}

\begin{figure}[hbtp]
  \centering
  \includegraphics[width=\columnwidth]{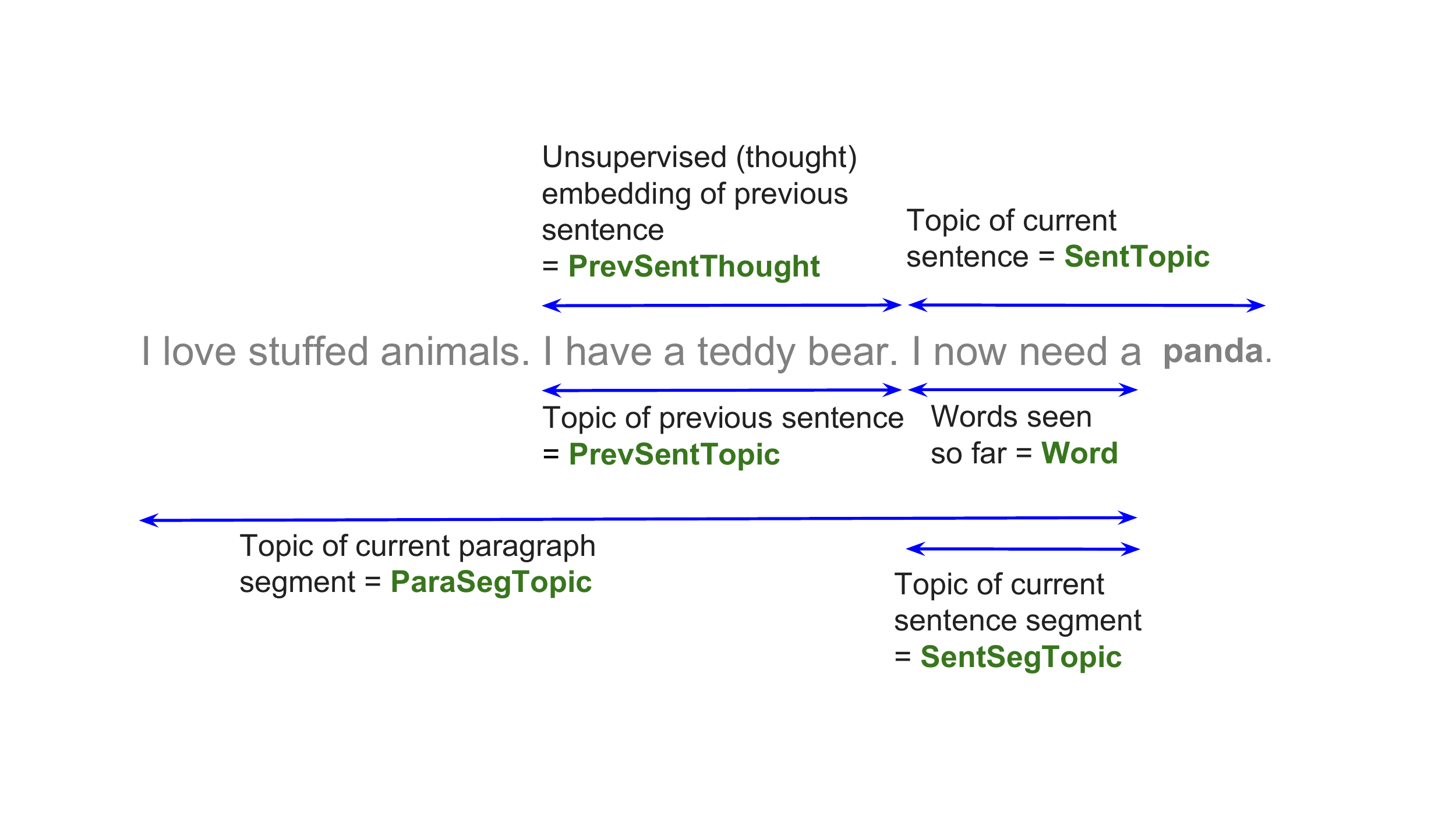}
  \caption{Hierarchical Features used in CLSTM models}
  \label{fig:features}
\end{figure}

\subsubsection{Features}
\label{sec:features}

We trained different types of CLSTM models for the word prediction
task.  The different types of features used in the different CLSTM
models are shown schematically in Figure~\ref{fig:features}. The
hierarchical features that we used in different variants of the word
prediction model are:

\begin{enumerate}

\item PrevSentTopic = TopicID of the topic computed based on all the words of the previous sentence, i.e., $T = Topic(w_{i-1,0} \ldots w_{i-1,n_{i-1}-1})$.
\item SentSegTopic = TopicID of the topic computed based on the words of the current sentence prefix until the current word, i.e., $T = Topic(w_{i,0} \ldots w_{i,j})$.
\item ParaSegTopic =  TopicID of the topic computed based on the paragraph prefix until the current word, i.e., $T = Topic(w_{0,0} \ldots w_{i,j})$. 

\end{enumerate}

where $T$ is defined in Section~\ref{sec:wordpred}.

% TODO: you say you are interested in maximizing: p(T_{i+1} | s_i,
% T_i) but your component model is p(w_{i,j+1} | w_{i,j}, T..). It’s
% not clear how you get from w to t given this component model. You
% will need some sort of prior on T, or a joint model that predicts
% w,t in order to compute the desired quantity.

\subsubsection{Datasets}
\label{sec:data}

For our experiments, we used the whole English corpus from Wikipedia
(snapshot from 2014/09/17). There were 4.7 million documents in the
Wikipedia dataset, which we randomly divided into 3 parts: 80\% was
used as train, 10\% as validation and 10\% as test set. Some relevant
statistics of the train, test and validation data sets of the
Wikipedia corpus are given in Table~\ref{tab:wikidata}.

\begin{table}[hbtp]
  \begin{center}
    \caption{Wikipedia Data Statistics (M=million)}
    \begin{tabular}{|r||r|r|r|}
      \hline
      Dataset & \#Para & \#Sent & \#Word \\
      \hline \hline
      Train (80\%) & 23M & 72M & 1400M \\
      Validation (10\%) & 2.9M & 8.9M & 177M \\
      Test (10\%) & 3M & 9M & 178M \\ \hline
    \end{tabular}
    \label{tab:wikidata}
  \end{center}
\end{table}

We created the vocabulary from the words in the training data, filtering
out words that occurred less than a particular threshold count in the
total dataset (threshold was 200 for Wikipedia). This resulted in a
vocabulary with 129K unique terms, giving us an out-of-vocabulary rate
of 3\% on the validation dataset.

For different types of text segments (e.g., segment, sentence,
paragraph) in the training data, we queried HTM and got the most
likely topic category. That gave us a total of $\approx$1600 topic
categories in the dataset.

\subsubsection{Results}
\label{wordpred-results}

\begin{table*}[hbtp]
  \begin{center}
    \caption{Test Set Perplexity for Word Prediction task}
    \begin{tabular}{|c||c|c|c|}
      \hline
      Input & Num Hidden & Num Hidden & Num Hidden \\
      Features & Units = 256 & Units = 512 & Units = 1024 \\
      \hline \hline
      Word & 38.56 & 32.04 & 27.66 \\
      Word + PrevSentTopic & 37.79 & 31.44 & 27.81 \\
      Word + SentSegTopic & 38.04 & 31.28 & 27.34 \\
      Word + ParaSegTopic & 38.02 & 31.41 & 27.30 \\
      Word + PrevSentTopic + SentSegTopic & 38.11 & 31.22 & 27.31 \\
      Word + SentSegTopic + ParaSegTopic & 37.65 & 31.02 & 27.10 \\ \hline
    \end{tabular}
    \label{tab:results-wordpred}
  \end{center}
\end{table*}

We trained different CLSTM models with different feature variants till
convergence, and evaluated their perplexity on the holdout test
data. Here are some key observations about the results (details in
Table~\ref{tab:results-wordpred}):\\

\noindent 1) The ``Word + SentSegTopic + ParaSegTopic'' CLSTM model is
the best model, getting the best perplexity.  This particular LSTM
model uses both sentence-level and paragraph-level topics as features,
implying that both local and long-range context is important for
getting the best performance. \\

\noindent 2) When current segment topic is present, the topic of the
previous sentence does not matter. \\

\noindent 3) As we increased the number of hidden units, the
performance started improving. However, beyond 1024 hidden units,
there were diminishing returns --- the gain in performance was
out-weighed by the substantial increase in computational overhead. \\

Note that we also trained a distributed n-gram model with ``stupid
backoff'' smoothing~\cite{Brants07} on the Wikipedia dataset, and it
gave a perplexity of $\approx$80 on the validation set. We did not
train a n-gram model with Knesner-Ney (KN) smoothing on the Wikipedia
data, but on the Google News data (from a particular snapshot) the KN
smoothed n-gram model gave a perplexity of 74 (using
5-grams).
%~\cite{vinyals_personal}.

Note that we were not able to compare our CLSTM models to other
existing techniques for integrating topic information into LSTM models
(e.g., Mikolov et al.~\cite{MikolovandZweigSLT-12}), since we didn't
have access to implementations of these approaches that can scale to
the vocabulary sizes ($\approx$ 100K) and dataset sizes we worked with
(e.g., English Wikipedia, Google News snapshot). Hence, we used a
finely-tuned LSTM model as a baseline, which we also trained at scale
on these datasets.

\section{Next Sentence Selection}
\label{sec:scoring}

\begin{figure}[hbtp]
  \centering
  \includegraphics[width=1.0\columnwidth]{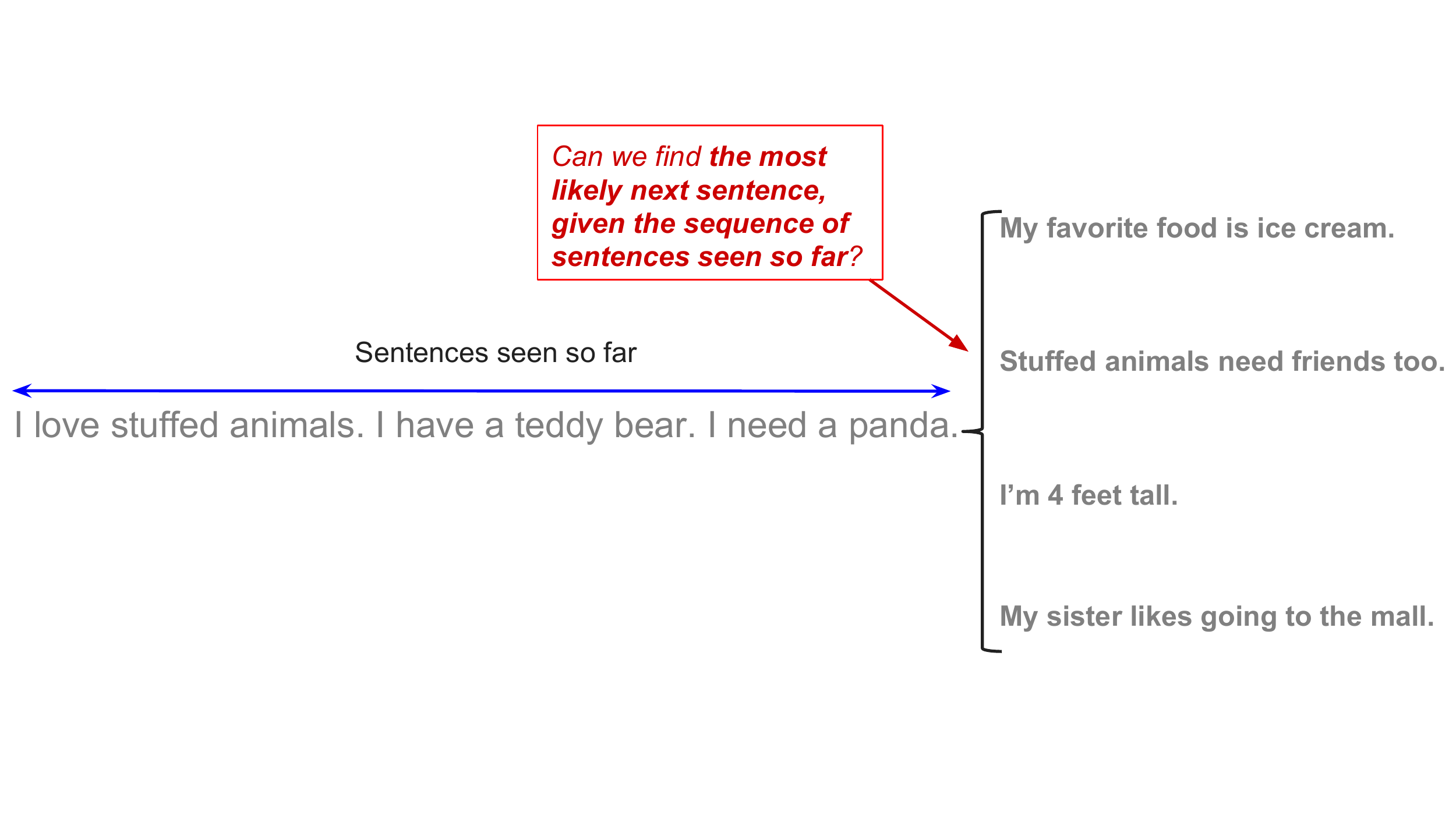}
  \caption{Next Sentence Selection Example}
  \label{fig:scoring}
\end{figure}

We next focus on the next sentence scoring task, where we are given a
sequence of sentences and the goal is to find the most probable next
sentence from a set of candidate sentences. 
An example of this task is shown in Figure~\ref{fig:scoring}. 
The task can be stated as follows:
given a model with parameters $\Theta$, a sequence of $p-1$ sentences
$s_0 \ldots s_{p-2}$ (with their corresponding topics $T_0 \ldots
T_{p-2}$), find the most likely next sentence $s_{p-1}$ from a candidate
set of next sentences $S$, such that:
\[
s_{p-1} = \arg\max_{s \in S} P(s | s_0 \ldots s_{p-2}, T_0 \ldots T_{p-2}, \Theta).
\]

\subsection{Problem Instantiation}
\label{sec:scoring-problem}

Suppose  we are given a set of sequences, where each sequence
consists of 4 sentences (i.e., we consider $p$=4). Let each sequence
be $S_i = <A_i B_i C_i D_i>$, and the set of sequences be
$\{S_1,\ldots,S_k\}$. Given the prefix $A_i B_i C_i$ of the sequence
$S_i$ as context (which we will denote to be $Context_i$), we consider
the task of correctly identifying the next sentence $D_i$ from a
candidate set of sentences: $\{D_0, D_1,\ldots, D_{k-1}\}$. For each
sequence $S_i$, we compute the accuracy of identifying the next
sentence correctly. The accuracy of the model in detecting the
correct next sentence is computed over the set of sequences
$\{S_1,\ldots,S_k\}$.

%% The schematic outline of the task is given in Figure~\ref{fig:scoring-problem}.
%% \begin{figure}[hbtp]
%%   \includegraphics[width=\columnwidth]{figs/fig-6.pdf}
%%   \caption{Next Sentence Scoring Problem}
%%   \label{fig:scoring-problem}
%% \end{figure}

\subsection{Approach}

We train LSTM and CLSTM models specifically for the next sentence
prediction task. Given the context $Context_i$, the
models find the $D_i$ among the set $\{ D_0 \ldots D_{k-1}\}$ that
gives the maximum (normalized) score, defined as follows:

\begin{eqnarray}
\forall i, score = \frac{P(D_i | Context_i)}{\frac{1}{k} \sum_{j=0}^{k-1} P(D_i | Context_j)}
\label{fig:hlstm-eqn}
\end{eqnarray}

\begin{figure}[hbtp]
  \includegraphics[width=\columnwidth]{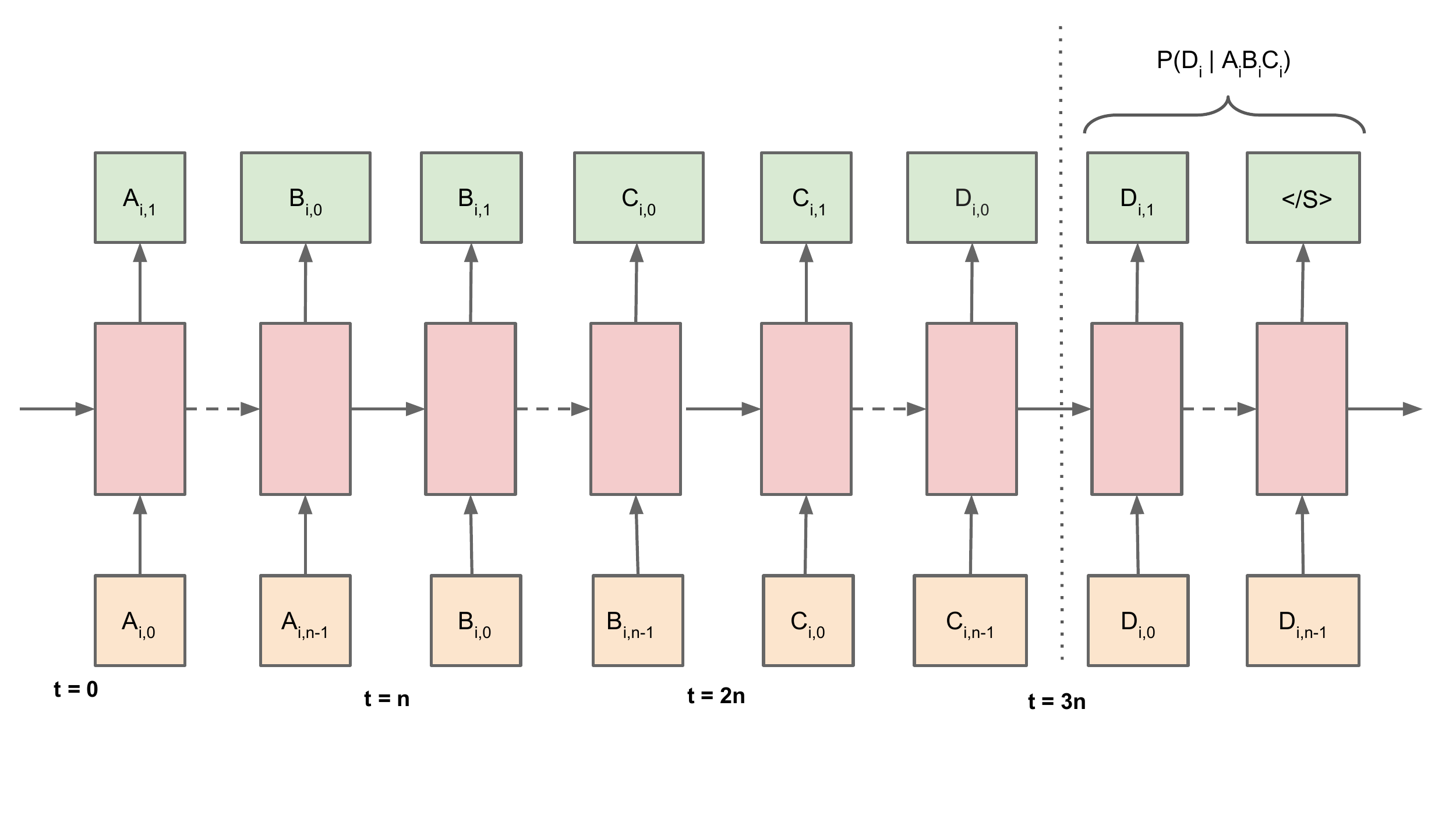}
  \caption{Next Sentence Scoring in CLSTM model}
  \label{fig:scoring-schematic}
\end{figure}

In the above score, the conditional probability terms are estimated
using inference on the LSTM and CLSTM models. In the numerator, the
probability of the word sequence in $D_i$, given the prefix context
$Context_i$, is estimated by running inference on a model whose state
is already seeded by the sequence $A_iB_iC_i$ (as shown in
Figure~\ref{fig:scoring-schematic}).
The normalizer term $\frac{1}{k} \sum_{j=0}^{k-1} P(D_i | Context_j)$
in the denominator of Equation~\ref{fig:hlstm-eqn} is the point
estimate of the marginal probability $P(D_i)$ computed over the set of
sequences, where the prior probability of each prefix context is
assumed equal, i.e., $P(Context_j) = \frac{1}{k}, j \in [0, k-1]$. The
normalizer term adjusts the score to account for the popularity of a
sentence $D_i$ that naturally has a high marginal probability $P(D_i)$
--- we do not allow the popularity of $D_i$ to lead to a high score.

%Instead, we want the score to reflect the relative
%improvement of predicting $D_i$, comparing the probability of
%$Context_i$ predicting $D_i$ to the average probability of a prefix
%context in the set predicting $D_i$.

Note that for task of next sentence scoring, it's ok to use words of
the next sentence when selecting the ``best'' next sentence. This is
because in the task, the possible alternatives are all provided to the
model, and the main goal of the model is scoring the alternatives and
selecting the best one. This setting is seen in some real-world
applications, e.g., predicting the best response to an email from a
set of candidate responses~\cite{smart_reply}.

\subsection{Model}
\label{sec:scoring-model}

%% \begin{figure}[hbtp]
%%   \centering
%%   \includegraphics[width=\columnwidth]{figs/fig-8.pdf}
%%   \caption{Schematic outline of Next Sentence Scoring Dataset}
%%   \label{fig:scoring-blocks}
%% \end{figure}

We trained a baseline LSTM model on the words of $A_i$, $B_i$ and
$C_i$ to predict the words of $D_i$. The CLSTM model uses words from
$A_i$, $B_i$, $C_i$, and topics of $A_i$, $B_i$, $C_i$ and $D_i$, to
predict the words of $D_i$. Note that in this case we can use the
topic of $D_i$ since all the candidate next sentences are given as
input in the next sentence scoring task.

For 1024 hidden units, the perplexity of the baseline LSTM model after
convergence of model training is 27.66, while the perplexity of the
CLSTM model at convergence is 24.81. This relative win of 10.3\% in an
intrinsic evaluation measure (like perplexity) was the basis for
confidence in expecting good performance when using this CLSTM model
for the next sentence scoring task.

\subsection{Experimental Results}
\label{sec:scoring-results}

We ran next sentence scoring experiments with a dataset generated from
the test set of the corpora.
%% Note that no models were trained or tuned
%% on this data, since the models were all trained on the training
%% dataset and evaluated on the test dataset (for perplexity estimation
%% at model convergence). 
We divide the test dataset into 100 non-overlapping subsets. To create
the dataset for next sentence scoring, we did the following: (a)
sample 50 sentence sequences $<A_i B_i C_i D_i>$ from 50 separate
paragraphs, randomly sampled from 1 subset of the test set -- we call
this a block; (b) consider 100 such blocks in the next sentence
scoring dataset. So,
%% Figure~\ref{fig:scoring-blocks} shows the schematic of the
%% dataset preparation for next sentence scoring. 
overall there are 5000 sentence sequences in the final dataset. For
each sequence prefix $A_i B_i C_i$, the model has to choose the best
next sentence $D_i$ from the competing set of next sentences.

\begin{table}[hbtp]
  \begin{center}
    \caption{Accuracy of CLSTM on next sentence scoring}
    \begin{tabular}{|c||c|c|}
      \hline
      LSTM & CLSTM & Accuracy Increase \\ 
      \hline \hline
      $52\% \pm 2\%$ & $63\% \pm 2\%$ & $21\% \pm 9\%$ \\ \hline
    \end{tabular}
    \label{tab:results-scoring}
  \end{center}
\end{table}

The average accuracy of the baseline LSTM model on this dataset is
$52\%$, while the average accuracy of the CLSTM model using word +
sentence-level topic features is $63\%$ (as shown in
Table~\ref{tab:results-scoring}). So the CLSTM model has an average
improvement of $21\%$ over the LSTM model on this dataset. Note that
on this task, the average accuracy of a random predictor that randomly
picks the next sentence from a set of candidate sentences would be
$2\%$.

We also ran other experiments, where the negatives (i.e., 49 other
sentences in the set of 50) were not chosen randomly --- in one case
we considered all the 50 sentences to come from the same HTM topic,
making the task of selecting the best sentence more difficult. In this
case, as expected, the gain from using the context in CLSTM was larger
--- the CLSTM model gave larger improvement over the baseline LSTM
model than in the case of having a random set of negatives.

\subsection{Error Analysis}
\label{sec:error-analysis}

\begin{figure*}[hbtp]
  \centering
  \fbox{\includegraphics[width=0.65\textwidth]{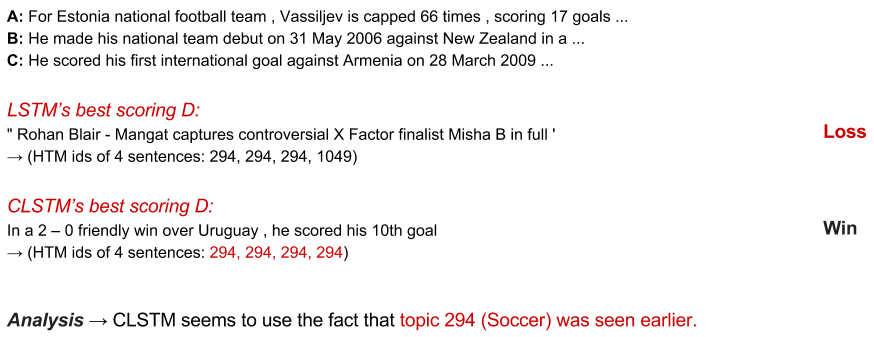}}
  \caption{Error Type A: CLSTM correct, LSTM incorrect}
  \label{fig:errorA}
\end{figure*}

\begin{figure*}[hbtp]
  \centering
  \fbox{\includegraphics[width=0.65\textwidth]{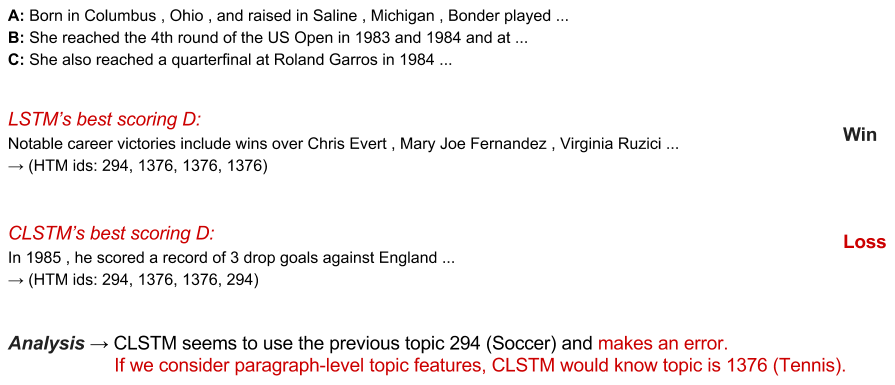}}
  \caption{Error Type B: CLSTM incorrect, LSTM correct}
  \label{fig:errorB}
\end{figure*}

\begin{figure*}[hbtp]
  \centering
  \fbox{\includegraphics[width=0.65\textwidth]{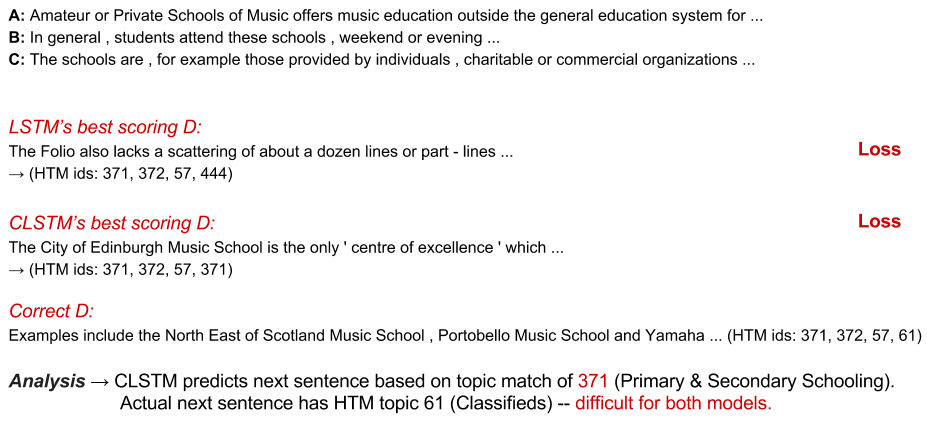}}
  \caption{Error Type C: CLSTM and LSTM both incorrect}
  \label{fig:errorC}
\end{figure*}

Figures~\ref{fig:errorA}-\ref{fig:errorC} analyze different types of
errors made by the LSTM and the CLSTM models, using samples drawn from
the test dataset.

\section{Sentence Topic Prediction}
\label{sec:topic}

%% \begin{figure}[hbtp]
%%   \centering
%%   \includegraphics[width=1.1\columnwidth]{figs/fig-9.pdf}
%%   \caption{Hierarchical Features used in Sentence Topic Prediction}
%%   \label{fig:sentence-pred-features}
%% \end{figure}

The final task we consider is the following: if we are given the words
and the topic of the current sentence, can we predict the topic of the
next sentence?  This is an interesting problem for dialog systems,
where we ask the question: given the utterance of a speaker, can we
predict the topic of their next utterance? This can be used in various
applications in dialog systems, e.g., intent modeling.

The sentence topic prediction problem can be formulated as follows:
given a model with parameters $\Theta$, words in the sentence $s_i$
and corresponding topic $T_i$, find the next sentence topic $T_{i+1}$
that maximizes the following probability -- $P(T_{i+1} | s_i, T_i,
\Theta)$. Note that in this case we train a model to predict the topic
target instead of the joint word/topic target, since we empirically
determined that training a model with a joint target gave lower
accuracy in predicting the topic compared to a model that only tries
to predict the topic as a target.

%% \begin{figure*}[bt]
%%   \centering
%%   \includegraphics[width=0.8\textwidth]{figs/fig-10.pdf}
%%   \caption{Unrolled Sentence Topic Prediction Architecture}
%%   \label{fig:topic-unrolled}
%% \end{figure*}

\subsection{Model}
\label{sec:topic-model}

For the sentence topic prediction task, we determined through ablation
experiments that the unrolled model architecture,
% (as shown in Figure~\ref{fig:topic-unrolled}), 
where each sentence in a paragraph
is modeled by a separate LSTM model, has better performance than the
rolled-up model architecture used for word prediction 
%(as shown in Figure~\ref{fig:clstm-distbelief}), 
where the sentences in a paragraph
are input to a single LSTM.

\subsection{Experiments}
\label{sec:topic-expt}

In our experiments we used the output of HTM as the topic of each
sentence. Ideally we would associate a ``supervised topic'' with each
sentence (e.g., the supervision provided by human raters). However,
due to the difficulty of getting such human ratings at scale, we used
the HTM model to find topics for the sentences. Note that the HTM
model is trained on human ratings.

We trained 2 baseline models on this dataset. The Word model uses the
words of the current sentence to predict the topic of the next
sentence -- it determines how well we can predict the topic of the
next sentence, given the words of the current sentence.  We also
trained another baseline model, SentTopic, which uses the sentence
topic of the current sentence to predict the topic of the next
sentence -- the performance of this model will give us an idea of the
inherent difficulty of the task of topic prediction. We trained a
CLSTM model (Word+SentTopic) that uses both words and topic of the
current sentence to predict the topic of the next
sentence. Figure~\ref{fig:features} shows the hierarchical features
used in the CLSTM model. We trained all models with different number
of hidden units: 256, 512, 1024. Each model was trained till
convergence. Table~\ref{tab:results-topic} shows the comparison of the
perplexity of the different models. The CLSTM model beats the baseline
SentTopic model by more than 12\%, showing that using hierarchical
features
% (word + sentence topic) 
is useful for the task of sentence topic prediction too.

\begin{table}[hbtp]
  \begin{center}
    \caption{Test Set Perplexity for sentence topic prediction (W=Word, ST=SentTopic)}
    \begin{tabular}{|c||c|c|c|}
      \hline
      Inputs & \#Hidden & \#Hidden & \#Hidden \\
             & units=256 & units=512 & units=1024 \\
      \hline \hline
      W & 24.50 & 23.63 & 23.29 \\
      ST & 2.75 & 2.75 & 2.76 \\
      W + ST & 2.43 & 2.41 & 2.43 \\ \hline
    \end{tabular}
    \label{tab:results-topic}
  \end{center}
\end{table}

\subsection{Comparison to BOW-DNN baseline}
\label{sec:bow-dnn}

%% \begin{table}[hbtp]
%%   \begin{center}
%%     \caption{Results of CLSTM model on current sentence topic prediction.}
%%     \begin{tabular}{|c|c|}
%%       \hline
%%       Num Hidden Units & CLSTM Perplexity \\
%%       \hline \hline
%%       256 & 15.9 \\
%%       512 & 15.6 \\
%%       1024 & 15.3 \\ \hline
%%     \end{tabular}
%%     \label{tab:bow-dnn}
%%   \end{center}
%% \end{table}

For the task of sentence topic prediction, we also compared the CLSTM
model to a Bag-of-Words Deep Neural Network (BOW-DNN)
baseline~\cite{Bai2014}. The BOW-DNN model extracts bag of words from
the input text, and a DNN layer is used to extract higher-level
features from the bag of words. For this experiment, the task setup we
considered was slightly different in order to facilitate more direct
comparison. The goal was to predict the topic of the next sentence,
{\em given words of the next sentence}. The BOW-DNN model was trained
only on word features, and got a test set perplexity of 16.5 on
predicting the sentence topic. The CLSTM model, trained on word and
topic-level features, got a perplexity of 15.3 on the same test set
using 1024 hidden units, thus outperforming the BOW-DNN model by
7.3\%.
%The detailed results of the CLSTM model in this experiment are shown in
%Table~\ref{tab:bow-dnn}.

\subsection{Using Unsupervised Topic Signals}
\label{sec:unsup}

\begin{figure*}[hbtp]
  \centering
  \includegraphics[width=0.75\textwidth]{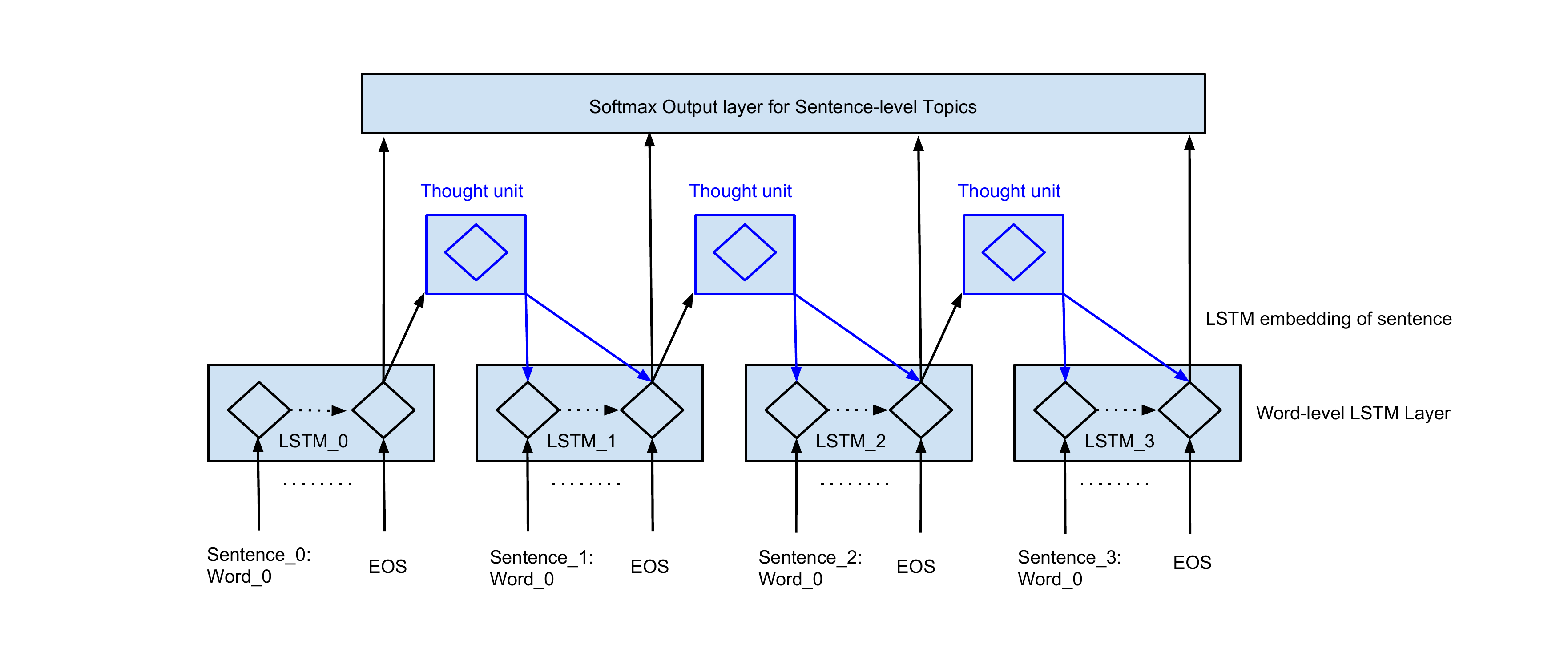}
  \caption{CLSTM model with Thought Vector}
  \label{fig:thought}
\end{figure*}

In our experiments with topic features, we have so far considered
supervised topic categories obtained from an extraneous source
(namely, HTM). One question arises: if we do not use extraneous topics
to summarize long-range context, would we get any improvements in
performance with unsupervised topic signals? To answer this question,
we experimented with ``thought embeddings'' that are intrinsically
generated from the previous context.  Here, the thought embedding from
the previous LSTM is used as the topic feature in the current LSTM (as
shown in Figure~\ref{fig:thought}), when making predictions of the
topic of the next sentence -- we call this context-based thought
embedding the ``thought vector''.\footnote{The term ``thought vector''
  was coined by Geoffrey Hinton~\cite{thought_vector}.}

In our approach, the thought vector inferred from the LSTM encoding of
sentence $n-1$ is used as a feature for the LSTM for sentence $n$, in
a recurrent fashion. Note that the LSTMs for each sentence in
Figure~\ref{fig:thought} are effectively connected into one long
chain, since we don't reset the hidden state at the end of each
sentence --- so the LSTM for the current sentence has access to the
LSTM state of the previous sentence (and hence indirectly to its
topic). But we found that directly adding the topic of the previous
sentence to all the LSTM cells of the current sentence is beneficial,
since it constraints all the current LSTM cells during training and
explicitly adds a bias to the model. Our experiments showed that it's
beneficial to denoise the thought vector signal using a
low-dimensional embedding, by adding roundoff-based projection.
%One key observation was that we 
%--- we empirically determined a dimensionality of 128 floats to give
%good test set perplexity. 
Initial experiments using thought vector for sentence-topic prediction
look promising. A CLSTM model that used word along with thought vector
(PrevSentThought feature in the model) from the previous sentence as
features gave a 3\% improvement in perplexity compared to a baseline
LSTM model that used only words as
features. Table~\ref{tab:results-thought} shows the detailed results.
%while Figure~\ref{fig:topic-unrolled} shows the CLSTM model
%architecture.

When we used thought vectors, our results improved over using a
word-only model but fell short of a CLSTM model that used both words
and context topics derived from HTM.
%One hypothesis for why there is
%such a gap between the supervised and unsupervised signal is because
%the current low-dimensional embedding technique we use is based on a 
%of the dimensionality difference between the HTM topic space vs the
%low-dimensional thought vectors (1600 unique values vs 128 floats, respectively).  
%We ran some initial experiments that showed that projecting the thought
%vector to a low-dimensional embedding could help us get better results
%by bridging this gap. 
In the future, we would like to do more extensive experiments using
better low-dimensional projections (e.g., using clustering or
bottleneck mechanisms), so that we can get comparable performance to
supervised topic modeling approaches like HTM.
% that have low dimensionality of the topic space.

Another point to note --- we have used HTM as a topic model in our
experiments as that was readily available to us. However, the CLSTM
model can also use other types of context topic vectors generated by
different kinds of topic modeling approaches, e.g., LDA, KMeans.

\section{Results on Google News data}
\label{sec:news}

We also ran experiments on a sample of documents taken from a recent
(2015/07/06) snapshot of the internal Google News English
corpus\footnote{Note that this snapshot from Google News is internal
  to Google, and is separate from the One Billion Word
  benchmark~\cite{ChelbaMSGBK13}.}. This subset had 4.3 million
documents, which we divided into train, test and validation
datasets. Some relevant statistics of the datasets are given in
Table~\ref{tab:newsdata}. We filtered out words that occurred less
than 100 times, giving us a vocabulary of 100K terms.

\begin{table}[hbtp]
  \begin{center}
    \caption{Test Set Perplexity for sentence topic prediction using Thought vector (W=Word, PST=PrevSentThought)}
    \begin{tabular}{|c||c|c|c|}
      \hline
      Inputs & \#Hidden & \#Hidden & \#Hidden \\
             & units=256 & units=512 & units=1024 \\
      \hline \hline
      W & 24.50 & 23.63 & 23.29 \\
      W + PST & 24.38 & 23.03 & 22.59 \\ \hline
    \end{tabular}
    \label{tab:results-thought}
  \end{center}
\end{table}

\begin{table}[hbtp]
  \begin{center}
    \caption{Statistics of Google News dataset (M=million)}
    \begin{tabular}{|r||r|r|r|}
      \hline
      Dataset & \#Para & \#Sent & \#Word \\
      \hline \hline
      Train (80\%) & 6.4M & 70.5M & 1300M \\
      Validation (10\%) & 0.8M & 8.8M & 169M \\
      Test (10\%) & 0.8M & 8.8M & 170M \\ \hline
    \end{tabular}
    \label{tab:newsdata}
  \end{center}
\end{table}

We trained the baseline LSTM and CLSTM models for the different tasks,
each having 1024 hidden units. Here are the key results: \\

1) {\bf Word prediction task:} LSTM using only words as features had
perplexity of $\approx$ 37. CLSTM improves on LSTM by $\approx$ 2\%, using words,
sentence segment topics and paragraph sentence topics.

2) {\bf Next sentence selection task:} LSTM gave an accuracy of $\approx$ 39\%. 
CLSTM had an accuracy of $\approx$ 46\%, giving a 18\% improvement on average.

3) {\bf Next sentence topic prediction task:} LSTM using only current 
sentence topic as feature gave perplexity of $\approx$ 5. CLSTM improves on LSTM by 
$\approx$ 9\%, using word and current sentence topic as features. \\

\noindent As we see, we get similar improvements of CLSTM model over LSTM model for both the 
Wikipedia and Google News datasets, for each of the chosen NLP tasks.

\section{Conclusions}
\label{sec:concl}

%% Our first approach was to consider a hierarchical model as shown in
%% Figure~\ref{fig:model1v-general}. Each level in this hierarchical LSTM
%% structure models the sequence of concepts in the corresponding level
%% of the document hierarchy --- the lowest level models the sequence of
%% words in a sentence, the next level models the sequence of sentences in
%% a paragraph, and so on. 

We have shown how using contextual features in a CLSTM model can be
beneficial for different NLP tasks like word prediction, next
sentence selection and topic prediction.
For the word prediction task CLSTM improves on state-of-the-art LSTM
by 2-3\% on perplexity, for the next sentence selection task CLSTM
improves on LSTM by $\approx$20\% on accuracy on average, while for
the topic prediction task CLSTM improves on state-of-the-art LSTM by
$\approx$10\% (and improves on BOW-DNN by $\approx$7\%). 
These gains are all quite significant and we get similar gains on the
Google News dataset (Section~\ref{sec:news}), which shows the
generalizability of our approach.
%All these results are on the Wikipedia (English) dataset --- 
%We also did initial
%experiments with ``thought vectors'' in the CLSTM, which look
%promising --- CLSTM gave $\approx$3\% gain over LSTM baseline for topic
%prediction.
Initial results using unsupervised topic signal using with
vectors, instead of supervised topic models, are promising.
The gains obtained by using the context in the CLSTM model has major
implications of performance improvements in multiple important NLP
applications, ranging from sentence completion, question/answering,
and paraphrase generation to different applications in dialog systems.

%Specifically, on the next sentence scoring task the CLSTM model
%using hierarchical features beats state-of-the-art LSTM by $>$20\%,
%while on the sentence topic prediction tasks the CLSTM model beats the
%LSTM model by $>$10\% (on the Wikipedia data). We also ran multiple
%ablation experiments, which gave us insights into model architecture,
%e.g., a rolled-up architecture is best for word-prediction, while an
%unrolled architecture (with no connection between sentence-level
%LSTMs) is best for sentence-topic prediction.

\section{Future Work}
\label{sec:future}

\begin{figure*}[hbtp]
  \centering
  \includegraphics[width=0.7\textwidth]{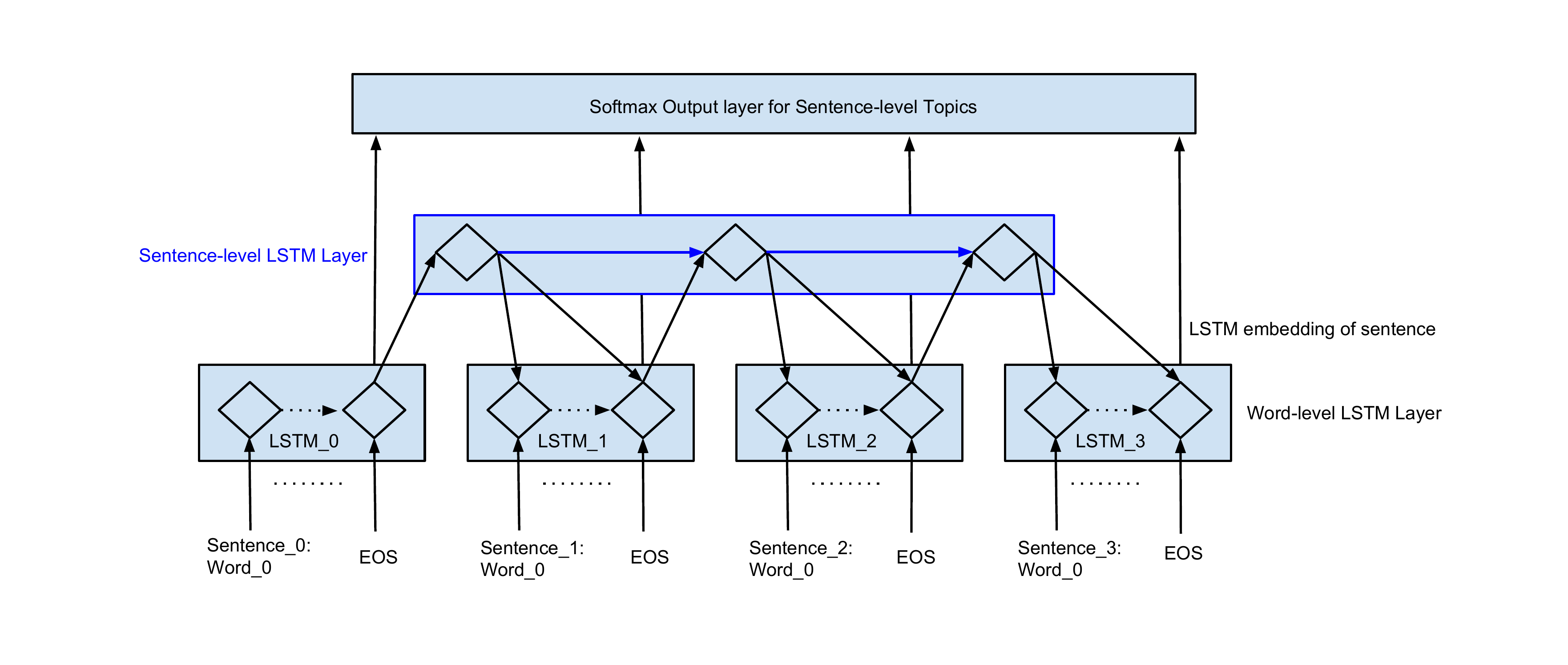}
  \caption{CLSTM model with Thought Vector and Sentence-level LSTM}
  \label{fig:model1v}
\end{figure*}

Our initial experiments on using unsupervised thought vectors for
capturing long-range context in CLSTM models gave promising results. A
natural extension of the thought vector model in
Figure~\ref{fig:thought} is a model that has a connection between the
hidden layers, to be able to model the ``continuity of
thought''. Figure~\ref{fig:model1v} shows one such hierarchical LSTM
(HLSTM) model, which has a 2-level hierarchy: a lower-level LSTM for
modeling the words in a sentence, and a higher-level LSTM for modeling
the sentences in a paragraph. The thought vector connection from the
LSTM cell in layer $n$ to the LSTM cells in layer $n-1$ (corresponding
to the next sentence) enables concepts from the previous context to be
propagated forward, enabling the ``thought'' vector of a sentence to
influence words of the next sentence. The connection between the
sentence-level hidden nodes also allows the model to capture the
continuity of thought. We would like to experiment with this model in
the future.

We would also like to explore the benefits of contextual features in
other applications of language modeling, e.g., generating better
paraphrases by using word and topic features.  Another interesting
application could be using topic-level signals in conversation
modeling, e.g., using Dialog Acts as a topic-level feature for next
utterance prediction.\\

\noindent {\bf Acknowledgments: } We would like to thank Ray Kurzweil,
Geoffrey Hinton, Dan Bikel, Lukasz Kaiser and Javier Snaider for
useful feedback on this work. We would also like to thank Louis Shao
and Yun-hsuan Sung for help in running some experiments.

\bibliographystyle{plain}
\bibliography{clstm,dean,vinyals}

\end{document}